\documentclass[conference]{IEEEtran}
\IEEEoverridecommandlockouts
\usepackage{cite}
\usepackage{amsmath,amssymb,amsfonts}
\usepackage{graphicx}
\usepackage{textcomp}
\usepackage{xcolor}
\usepackage{algorithm}
\usepackage{amsmath,amssymb,amsfonts}
\usepackage{newunicodechar}
\usepackage{adjustbox}
\usepackage{booktabs}
\usepackage{multirow}
\usepackage{xspace}
\usepackage{amssymb}
\usepackage{algpseudocode}
\usepackage{graphicx}
\usepackage{color,soul}
\usepackage{amsmath}
\usepackage{multirow}    % for multirow cells in tables
\usepackage{booktabs}    % for \toprule, \midrule, \bottomrule
\usepackage{array}   
\usepackage{tabularx}
\usepackage{enumitem}
\usepackage[hidelinks]{hyperref}

\def\BibTeX{{\rm B\kern-.05em{\sc i\kern-.025em b}\kern-.08em
    T\kern-.1667em\lower.7ex\hbox{E}\kern-.125emX}}
\begin{document}

\newcommand{\model}{\textsc{ImpuGAN}\xspace}

\title{\model: Learning Conditional Generative Models for Robust Data Imputation}

\author{

\IEEEauthorblockN{Zalish Mahmud}
\IEEEauthorblockA{\textit{Computer Science} \\
\textit{The University of Texas at El Paso}\\
El Paso, US \\
zmahmud@miners.utep.edu}
\and
\IEEEauthorblockN{Anantaa Kotal}
\IEEEauthorblockA{\textit{Computer Science} \\
\textit{The University of Texas at El Paso}\\
El Paso, US \\
akotal@utep.edu}
\and
\IEEEauthorblockN{Aritran Piplai}
\IEEEauthorblockA{\textit{Computer Science} \\
\textit{The University of Texas at El Paso}\\
El Paso, US \\
apiplai@utep.edu}
}

\maketitle

\begin{abstract}
Incomplete data are common in real-world applications. Sensors fail, records are inconsistent, and datasets collected from different sources often differ in scale, sampling rate, and quality. These differences create missing values that make it difficult to combine data and build reliable models. Standard imputation methods such as regression models, expectation–maximization, and multiple imputation rely on strong assumptions about linearity and independence. These assumptions rarely hold for complex or heterogeneous data, which can lead to biased or over-smoothed estimates. We propose \model, a conditional Generative Adversarial Network (cGAN) for imputing missing values and integrating heterogeneous datasets. The model is trained on complete samples to learn how missing variables depend on observed ones. During inference, the generator reconstructs missing entries from available features, and the discriminator enforces realism by distinguishing true from imputed data. This adversarial process allows \model to capture nonlinear and multimodal relationships that conventional methods cannot represent. In experiments on benchmark datasets and a multi-source integration task, \model achieves up to 82\% lower Earth Mover’s Distance (EMD) and 70\% lower mutual-information deviation (MI) compared to leading baselines. These results show that adversarially trained generative models provide a scalable and principled approach for imputing and merging incomplete, heterogeneous data. Our model is available at: \href{https://github.com/zalishmahmud/impuganBigData2025}{\texttt{github.com/zalishmahmud/impuganBigData2025}}.

\end{abstract}


\section{Introduction}

The rapid growth of data collection in domains such as healthcare, finance, environmental monitoring, and the social sciences has created new opportunities for data-driven research. However, the value of these datasets is often reduced by the widespread problem of missing data. Missing entries may arise from equipment failures, inconsistent reporting, privacy restrictions, or human error. They are rarely random and can introduce significant bias in statistical analyses and predictive models. The problem becomes even more difficult when data from multiple sources must be combined. These heterogeneous datasets often differ in sampling rate, measurement scale, granularity, and noise characteristics. In such settings, missing values are not isolated gaps but systematic issues that complicate data fusion and analysis.

Traditional methods provide useful but limited solutions. Simple heuristics, such as mean or median substitution, are efficient but ignore dependencies among variables and often distort variance. Statistical models, including expectation–maximization (EM) \cite{dempster1977maximum} and multiple imputation by chained equations (MICE) \cite{vanbuuren2011mice}, rely on assumptions of linearity, multivariate normality, or independence that rarely hold in real-world data. These limitations are especially evident in high-dimensional or multi-source contexts, where relationships are nonlinear and multimodal. Machine learning approaches, such as k-nearest neighbors (kNN) and tree-based methods like MissForest, improve flexibility but still struggle to capture complex joint distributions across diverse data sources \cite{troyanskaya2001missing}.

Recent advances in deep learning have expanded the range of tools for data imputation. Autoencoders and variational autoencoders (VAEs) learn latent representations that can reconstruct missing entries in a data-driven manner. However, these models typically minimize reconstruction error without ensuring that the reconstructed data follow the true underlying distribution, resulting in imputations that may be plausible numerically but not statistically representative. Generative Adversarial Network (GAN)–based approaches, such as the Generative Adversarial Imputation Net (GAIN) \cite{yoon2018gain}, address this issue by using adversarial training to enforce realism in imputed values. Yet most existing GAN frameworks are designed for single-source datasets and do not address the additional challenges of multi-source integration, where missingness interacts with heterogeneity, noise, and alignment issues. These limitations motivate three key research questions:
\begin{enumerate}[leftmargin=*, noitemsep]
\item \textbf{How can generative models be extended to perform imputation on heterogeneous, multi-source data?} Different sources introduce alignment and dependency constraints that must be preserved during imputation.
\item \textbf{How can imputation methods maintain statistical dependencies between observed and missing variables?} Simple substitutions often distort correlations across sources, reducing downstream model reliability.
\item \textbf{Can adversarial training provide scalable and realistic imputations that outperform existing methods across domains and missingness mechanisms?} We investigate whether conditional adversarial learning can more accurately capture complex conditional distributions than traditional or machine learning–based approaches.
\end{enumerate}

To address these questions, we propose \model, a novel framework for data imputation and integration based on {conditional Generative Adversarial Networks (cGANs)}. The key insight is to train the model on complete datasets, learning the conditional distributions of missing values given observed data. During inference, the generator reconstructs missing entries by conditioning on the available features, while the discriminator evaluates whether the completed data vectors are indistinguishable from real, fully observed samples. This adversarial interplay encourages imputations that are not only plausible on an entry-by-entry basis but also consistent with the global data distribution. Importantly, \model is designed to handle heterogeneous sources, allowing it to merge datasets with varying structures and noise levels.  

Our contributions are threefold:
\begin{itemize}[leftmargin=*, noitemsep]
    \item \textbf{A novel adversarial imputation framework for heterogeneous datasets.} We introduce \model, which extends cGANs to the task of imputing missing values in multi-source contexts, ensuring that reconstructions respect both local conditioning and global consistency.  
    \item \textbf{A systematic comparison with baseline imputation methods.} We benchmark \model against simple heuristics based and statistical imputations methods as well as SOTA machine learning methods for data imputation. This comprehensive evaluation highlights the advantages of adversarial conditioning in capturing joint distributions.  
\item \textbf{Extensive empirical validation.} Using multiple benchmark datasets and a comprehensive array of evaluation methods, we show that \model achieves state-of-the-art performance in data imputation. In particular, it reduces distributional error (EMD) by up to \textbf{82\%} compared to GAIN on the Adult and Heart Disease datasets, while also improving correlation preservation (MI deviation) by up to \textbf{70\%}.

    \item \textbf{Generalizable insights into data fusion.} Beyond imputation accuracy, we demonstrate that \model facilitates downstream tasks such as classification and regression on fused datasets, improving performance relative to traditional preprocessing pipelines.  
\end{itemize}

By addressing the dual challenges of missing data and heterogeneous source integration, \model offers a scalable and principled alternative to existing imputation techniques. Our results suggest that adversarially trained generative models can significantly improve data quality, enhance interpretability, and unlock the full potential of multi-source analytics.  

\section{Related Work}

\subsection{Traditional Statistics–Based Imputation}
Classical statistical theory distinguishes three canonical missingness mechanisms: \textit{Missing Completely at Random (MCAR)}, \textit{Missing at Random (MAR)}, and \textit{Missing Not at Random (MNAR)}. These frameworks underpin principled inference under incomplete data \cite{10.1093/biomet/63.3.581,little2002statistical}. The \textit{Expectation–Maximization (EM)} algorithm provides a general maximum-likelihood estimation procedure for datasets with missing entries \cite{dempster1977maximum}. Multiple imputation, most commonly implemented via \textit{Multiple Imputation by Chained Equations (MICE)}, remains a widely used practical tool across domains \cite{vanbuuren2011mice,vanbuuren2007multiple,buuren2018flexible}. Hot-deck and donor-based methods have long been popular in survey statistics due to their intuitive design \cite{andridge2010review}.  
In addition to these classical approaches, heuristic methods such as \textit{k-Nearest Neighbors (kNN)} imputation and tree/forest-based techniques (e.g., MissForest) trade parametric assumptions for robustness, particularly in mixed-type data settings \cite{troyanskaya2001missing,stekhoven2012missforest}. Low-rank matrix completion techniques, such as nuclear-norm minimization in “softImpute,” exploit approximate low-dimensional structure and provide recovery guarantees under idealized assumptions \cite{mazumder2010spectral,candes2009exact}.
 
\subsection{Deep Neural Network–Based Imputation}
Deep neural architectures have been widely explored for data imputation. Early work on autoencoders learned compact latent representations capable of reconstructing inputs even when partially corrupted \cite{hinton2006reducing,vincent2010stacked}. Denoising autoencoders inspired methods such as MIDA, a general-purpose imputation framework for tabular data \cite{gondara2018mida}.  
\textit{Variational Autoencoders (VAEs)} extended this paradigm by providing probabilistic latent-variable models with principled uncertainty estimates. Methods such as MIWAE (Missing-data Importance Weighted AutoEncoder), VAEAC (Variational Autoencoder with Arbitrary Conditioning), and GP-VAE (Gaussian Process Variational Autoencoder for time series) adapt the VAE framework to missing-data scenarios \cite{kingma2014vae,mattei2019miwae,ivanov2018vaeac,fortuin2020gpvae}.  
For sequential data, recurrent architectures such as \textit{Gated Recurrent Unit with Decay (GRU-D)} encode missingness patterns as informative signals \cite{che2018grud}, while \textit{Bidirectional Recurrent Imputation for Time Series (BRITS)} enforces consistency across forward and backward imputations \cite{cao2018brits}. More recently, transformer-style architectures such as SAITS (\textit{Self-Attention–based Imputation for Time Series}) have shown strong performance in capturing long-range temporal dependencies \cite{DU2023119619}.

\subsection{Generative Models: GANs and Conditional Generators}
\textit{Generative Adversarial Networks (GANs)} learn data distributions through an adversarial game between a generator and a discriminator \cite{goodfellow2014gan}. Training stability and diversity are enhanced by variants such as \textit{Wasserstein GAN with Gradient Penalty (WGAN-GP)} \cite{gulrajani2017wgangp} and \textit{Packaged GAN (PacGAN)} \cite{lin2018pacgan}. \textit{Conditional GANs (cGANs)} extend this paradigm by conditioning the generator on side information, enabling controllable and context-aware data generation \cite{mirza2014cgan, kotal2024differentially}. CTGAN introduced training-by-sampling strategies and mode-specific normalization to better handle mixed discrete/continuous variables and imbalanced categories \cite{xu2019modeling}, that has been extended to applications in security, healthcare, finance etc. \cite{kotal2022privetab}. 

\subsection{Generative-Model–Based Imputation}
Several methods directly apply generative models to missing data imputation. \textit{Generative Adversarial Imputation Networks (GAIN)} recast the imputation problem as a mask-discrimination task with a hint mechanism to guide training \cite{yoon2018gain}. MisGAN jointly models both the data distribution and the missingness process, allowing learning directly from incomplete datasets \cite{li2019misgan}. Beyond tabular data, adversarial inpainting methods in computer vision, such as Context Encoders \cite{pathak2016context}, inspired conditional synthesis strategies that influenced non-vision imputers. In the temporal domain, \textit{TimeGAN} combines adversarial training with sequence modeling to generate realistic and temporally consistent imputations \cite{yoon2019timegan}.  Non-GAN generative approaches also contribute: MIWAE, VAEAC, and GP-VAE provide probabilistic imputations, while diffusion-based models such as CSDI (\textit{Conditional Score-based Diffusion for Imputation}) and sequence models like NAOMI (\textit{Non-Autoregressive Multiresolution Imputation}) highlight the growing importance of uncertainty-aware and distributionally faithful imputations \cite{mattei2019miwae,ivanov2018vaeac}.

While traditional statistical methods provide useful baselines, they often rely on restrictive linearity or independence assumptions and fail to scale in high-dimensional heterogeneous settings. Deep learning approaches such as autoencoders and recurrent models improve flexibility but primarily optimize reconstruction error, which can produce imputations that are numerically plausible yet not distributionally faithful. Generative models, particularly GAN-based methods, represent a major step forward by enforcing realism through adversarial training. However, existing frameworks such as GAIN and MisGAN are typically designed for single-source datasets and struggle with the challenges of integrating heterogeneous data sources that differ in structure, granularity, and noise characteristics.  

\model addresses this critical limitation by extending conditional GANs to the setting of heterogeneous, multi-source data imputation. Through a multi-conditional adversarial mechanism, \model not only reconstructs missing entries in a distributionally consistent manner but also preserves correlations across disparate sources. This design enables imputations that are both realistic and coherent across heterogeneous domains—an aspect that prior methods have not explicitly targeted. In doing so, \model bridges the gap between distributional fidelity and cross-source integration, offering a scalable and principled solution for real-world incomplete data fusion.

\section{Methodology}
\label{sec_methodology}

\subsection{Problem Statement}

Let $\mathbf{X} \in \mathbb{R}^{n \times d}$ denote a dataset consisting of $n$ samples and $d$ features. In practice, many entries of $\mathbf{X}$ are missing due to incomplete data collection, sensor failures, or inconsistent reporting across heterogeneous sources. We represent the observed data by $\mathbf{X}_{\text{obs}}$ and the missing entries by $\mathbf{X}_{\text{miss}}$. A binary mask matrix $\mathbf{M} \in \{0,1\}^{n \times d}$ is introduced, where $M_{ij}=1$ if the $(i,j)$-th entry is observed and $M_{ij}=0$ otherwise. Thus, each sample can be expressed as:  

\[
\mathbf{x}_i = \mathbf{x}_{i,\text{obs}} \cup \mathbf{x}_{i,\text{miss}}, \quad i = 1, \ldots, n.
\]

The objective of \textit{data imputation} is to estimate the distribution of missing values conditioned on the observed entries:  

\[
p(\mathbf{X}_{\text{miss}} \mid \mathbf{X}_{\text{obs}}, \mathbf{M}).
\]

Classical approaches attempt to approximate this conditional distribution through statistical modeling or heuristic substitution. However, these methods often fail to capture complex nonlinear dependencies, multi-modal relationships, and correlations that exist across heterogeneous sources. Moreover, when datasets with different levels of granularity or noise must be merged, the imputation task becomes one of not only filling in missing entries but also ensuring coherence and consistency across disparate feature spaces.  

Formally, given a set of heterogeneous sources $\{\mathbf{X}^{(1)}, \mathbf{X}^{(2)}, \ldots, \mathbf{X}^{(s)}\}$, each with potentially incomplete observations, the goal is to learn a unified imputation model $\mathcal{I}(\cdot)$ such that:  

\[
\hat{\mathbf{X}} = \mathcal{I}\big(\mathbf{X}_{\text{obs}}, \mathbf{M}\big),
\]

where $\hat{\mathbf{X}}$ denotes the reconstructed dataset with missing values imputed in a manner that (i) preserves statistical fidelity with the underlying data distribution, (ii) maintains correlations across features and sources, and (iii) enables downstream learning tasks such as prediction or clustering.  

This formulation motivates the use of generative modeling techniques that explicitly approximate the conditional distribution $p(\mathbf{X}_{\text{miss}} \mid \mathbf{X}_{\text{obs}})$ in a data-driven way. In particular, we propose leveraging the expressive power of conditional Generative Adversarial Networks to approximate this distribution and produce imputations that are both realistic and distributionally consistent.  

\subsection{Why GANs Foster Diversity in Imputation}

A key challenge in missing data imputation is avoiding \textit{mode collapse} into overly simplistic or deterministic imputations. Many classical approaches, such as mean or regression-based imputation, generate a single ``best guess'' for each missing value. While this can minimize average reconstruction error, it ignores the inherent stochasticity of the underlying data distribution, leading to imputations that are biased, lack variability, and under-represent the true uncertainty.  

Generative Adversarial Networks (GANs) naturally foster diversity because they are trained to approximate the entire data distribution rather than a single point estimate. Formally, let $p_{\text{data}}(\mathbf{x})$ denote the distribution of complete data and let $p_{g}(\mathbf{x})$ denote the distribution induced by the generator $G(\cdot)$. The GAN objective is a two-player minimax game between the generator $G$ and discriminator $D$:  

\begin{align*}
    \min_{G} \max_{D} \; \mathbb{E}_{\mathbf{x} \sim p_{\text{data}}} \big[ \log D(\mathbf{x}) \big] 
+ \mathbb{E}_{\mathbf{z} \sim p_{\mathbf{z}}} \big[ \log (1 - D(G(\mathbf{z}))) \big]
\end{align*}

where $\mathbf{z}$ is sampled from a noise distribution $p_{\mathbf{z}}$ (e.g., Gaussian or uniform). The noise vector $\mathbf{z}$ introduces stochasticity into the generation process, ensuring that multiple plausible outputs can be sampled for the same conditioning input.  

\iffalse
For conditional GANs (cGANs), which are central to \model, the generator is conditioned on observed entries $\mathbf{x}_{\text{obs}}$ and the mask $\mathbf{M}$, yielding:  

\begin{align*}
    \min_{G} \max_{D} \; \mathbb{E}_{\mathbf{x}_{\text{obs}}, \mathbf{x}_{\text{miss}} \sim p_{\text{data}}}
\big[ \log D(\mathbf{x}_{\text{obs}},\mathbf{x}_{\text{miss}}) \big] +
\\ \mathbb{E}_{\mathbf{x}_{\text{obs}} \sim p_{\text{data}}, \mathbf{z} \sim p_{\mathbf{z}}} 
\big[ \log \big(1 - D(\mathbf{x}_{\text{obs}}, G(\mathbf{x}_{\text{obs}}, \mathbf{z}, \mathbf{M})) \big) \big]
\end{align*}

Here, the generator $G$ learns to produce missing values $\hat{\mathbf{x}}_{\text{miss}}$ conditioned on $\mathbf{x}_{\text{obs}}$, while the discriminator $D$ attempts to distinguish between true pairs $(\mathbf{x}_{\text{obs}}, \mathbf{x}_{\text{miss}})$ and imputed pairs $(\mathbf{x}_{\text{obs}}, \hat{\mathbf{x}}_{\text{miss}})$.  
\fi 

Because the generator samples from the latent noise $\mathbf{z}$, each run can yield a \textit{different but plausible} imputation for the same missing entries. This diversity is further reinforced by the adversarial training objective, which drives $p_g(\mathbf{x})$ to match $p_{\text{data}}(\mathbf{x})$ in distribution, rather than collapsing onto a single deterministic estimate. In contrast to mean- or regression-based imputations, GAN-based imputation preserves both the variability and multi-modality of the underlying data, producing imputations that are faithful not only in expectation but across the full data distribution.  

This property is particularly advantageous in multi-source integration, where missing entries may correspond to different but correlated phenomena across datasets. By capturing the joint distribution of observed and missing variables, GANs ensure that imputations respect source-level correlations while maintaining diversity across possible completions.  

\subsection{Implementation of \model}

\model is built specifically for imputing missing values in heterogeneous tabular data. It addresses two persistent challenges in existing generative approaches:  
(i) producing imputations that are both realistic and \emph{consistent with all observed attributes}, and  
(ii) handling both continuous and categorical variables in a way that respects multi-modality and imbalance.  

To tackle these, \model combines a multi-conditional adversarial framework with a tailored optimization scheme and a joint conditioning mechanism. Together, these ensure that imputations are faithful to the data distribution while remaining consistent with any requested conditions.  

\subsubsection{Optimization}

The learning process revolves around two neural networks: a generator $G$ and a discriminator $D$.  
- The generator proposes candidate values for the missing entries based on the observed ones, the observation mask $\mathbf{M}$, and injected stochastic noise $\mathbf{z}$.  
- The discriminator then judges whether the completed sample looks like a real, fully observed row or a synthetic one.  

To stabilize training, we use the \emph{PAC discriminator} strategy, where the discriminator evaluates groups of samples jointly, reducing variance in its gradients. The overall adversarial game is:

\begin{align*}
\min_{G} \max_{D} \;\;
\mathbb{E}_{\mathbf{x}\sim p_{\text{data}}}\!\left[\log D(\mathbf{x})\right]
\\+ \mathbb{E}_{\mathbf{z},\,\mathbf{x}_{\text{obs}},\,\mathbf{M}}\!\left[\log\left(1 - D(\mathbf{x}_{\text{obs}},G(\mathbf{x}_{\text{obs}},\mathbf{M},\mathbf{z}))\right)\right]
\end{align*}

To enforce consistency with user-specified conditions (e.g., categorical attributes that must remain fixed), we add a \textbf{conditional fidelity loss}. This loss penalizes the generator whenever its outputs deviate from the requested categories. The generator’s total loss is:

\[
\mathcal{L}_{G} = -\mathbb{E}_{\mathbf{z}}[D(\hat{\mathbf{x}})] 
+ \lambda_{\text{cond}} \mathcal{L}_{\text{cond}}.
\]

Training alternates between updating the discriminator and the generator, while applying gradient penalties and PAC discrimination for stability. This ensures that \model does not collapse to trivial averages but instead learns the full conditional distribution of missing values.  
\begin{figure}[h]
    \centering
    \includegraphics[width=0.9\linewidth]{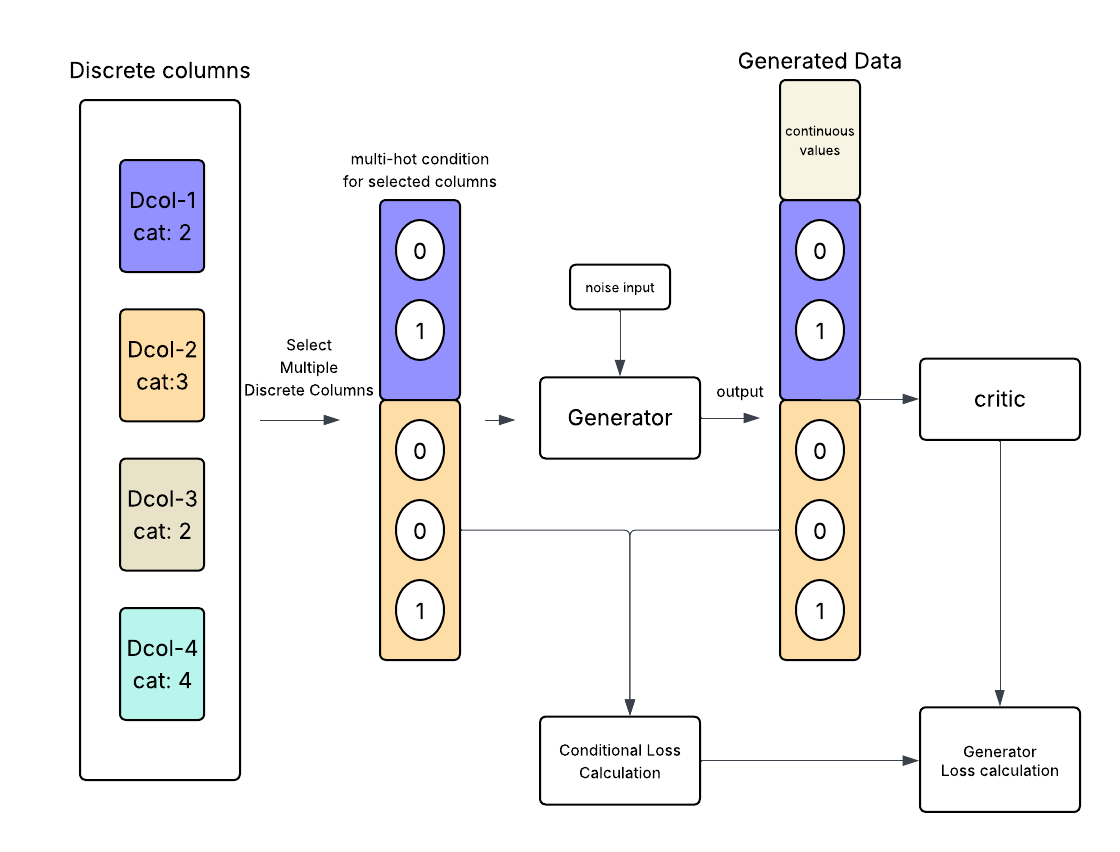}
    \caption{Conditional modeling in \model. 
    The multi-conditional vector allows the model to incorporate multiple observed categorical attributes simultaneously, 
    ensuring imputations respect all given conditions.}
    \label{fig:conditional_model}
\end{figure}
\subsubsection{Conditional Modeling via Joint Conditioning} 
As shown in Figure~\ref{fig:conditional_model}, A core contribution of \model is its \textbf{multi-conditional mechanism}. Unlike earlier approaches that condition on only one attribute, \model can incorporate \emph{any subset of observed categorical attributes} simultaneously.  

Formally, given the set $\mathcal{S}_{\text{disc}}$ of observed categorical attributes, we form a multi-hot condition vector: 

$\mathbf{cond} = \bigoplus_{i \in \mathcal{S}_{\text{disc}}}\mathbf{m}_i^{*} $; 

where $\mathbf{m}_i^{*}$ is the one-hot vector for the observed category in column $i$. The conditional loss then aggregates the cross entropy loss (CE) across all requested attributes:
\[
\mathcal{L}_{\text{cond}} = 
\sum_{i\in\mathcal{S}_{\text{disc}}}\!\text{CE}(\mathbf{m}_i^{*},\,\hat{\mathbf{d}}_i).
\]

This guarantees that imputations are not only plausible with respect to the overall data distribution but also strictly consistent with \emph{every requested categorical attribute}. Continuous features, by contrast, are modeled implicitly by the generator, reflecting their smoother and distributional nature.  
\subsubsection{Hard Sampling via Logit Override.}  
To encourage sharper categorical predictions, \model employs a hard sampling strategy. 
Instead of relying solely on soft probabilities, the highest-scoring category is occasionally selected directly 
by overriding the generator’s logits. 
This simple mechanism reduces uncertainty in categorical outputs and prevents mode collapse, 
while still allowing the model to explore alternative categories during training.

\begin{algorithm}[t]
\caption{\model Framework: Multi-Conditional Adversarial Training with Hard-Conditioned Sampling}
\label{alg:unified}
\begin{algorithmic}[1]
\Require data $X$, discrete cols $D$, batch size $B$, epochs $T$, critic steps $n_c$, lrs $\eta_G,\eta_D$, GP weight $\lambda$, cond.~weight $\alpha$, PAC size $p$, (optional) user request $(n,c^\ast)$
\Ensure trained $G$, transformer $\mathcal{T}$, sampler $\mathcal{S}$, and (if requested) synthetic $\hat X$
\State Transform $X$ with $\mathcal{T}$; build sampler $\mathcal{S}$
\State Init $G,D$ with Adam$(\eta_G,\eta_D)$
\For{$t=1..T$}
  \For{batches $s$}
    \For{$k=1..n_c$} 
      \State $c\gets\textsc{SampleCond}(B)$; 
      \State $x_r\gets\textsc{SampleReal}(X,B,c)$
      \State $x_f\gets G(z,c)$, $z\sim\mathcal{N}(0,I)$
      \State $L_D\gets -\mathbb{E}[D(x_r,c)]+\mathbb{E}[D(x_f,c)]+\textsc{GP}$
      \State Update $D$
    \EndFor
    \State $c\gets\textsc{SampleCond}(B)$; $x_f\gets G(z,c)$
    \State $L_G\gets -\mathbb{E}[D(x_f,c)]+\alpha\,\textsc{CondLoss}(x_f,c)$
    \State Update $G$
  \EndFor
\EndFor
\If{sampling $(n,c^\ast)$ requested}
  \While{$|\tilde X| < n$}
  \State sample $z \sim \mathcal{N}(0,I)$
  \State $x \gets \textsc{Activate}(G([z;c^\ast]))$ 
  \State $x \gets \textsc{HardApply}(x, c^\ast)$ 
  \State append $x$ to $\hat X$
\EndWhile
\EndIf
\State \Return $G,\mathcal{T},\mathcal{S}$ (and $\hat X$ if requested)
\end{algorithmic}
\end{algorithm}

% ----- Subroutine: HardApply on discrete spans -----
\begin{algorithm}[t]
\caption{\textsc{HardApply}$(x, c^\ast)$}
\label{alg:hard-apply}
\begin{algorithmic}[1]
\Require activated vector $x$ (with softmax spans), condition $c^\ast$ (col $\to$ category)
\Ensure $x$ with conditioned spans set exactly
\For{each conditioned column $j$ with desired category $q$}
  \State find the softmax span for column $j$
  \State set that span to one-hot: entry $q \gets 1$, others $\gets 0$
\EndFor
\State \Return $x$
\end{algorithmic}
\end{algorithm}

\section{Implementation}
\subsection{Datasets}
To evaluate the effectiveness of the \model framework, we conducted experiments on three publicly available datasets spanning demography and healthcare. These datasets were selected for their mix of continuous and categorical variables and their widespread use in benchmarking classification models.

\begin{itemize}
    \item \textbf{Adult (Census Income) Dataset} \cite{becker1996adult}: \textbf{Instances:} 48{,}842; \textbf{Predictors:} 14 \,(6 continuous, 8 categorical). Records include demographic and economic attributes such as age, education, occupation, and native country. The task is to predict whether annual income exceeds \$50K.

    \item \textbf{PIMA Indians Diabetes Dataset} \cite{pima_diabetes_niddk}: \textbf{Instances:} 768; \textbf{Predictors:} 6 \,(6 continuous, 2 categorical). Clinical measurements (e.g., glucose, blood pressure, BMI, age) from female Pima Indian patients are used to predict diabetes onset.

    \item \textbf{Heart Disease (UCI, Cleveland) Dataset} \cite{heart_disease_45}: \textbf{Instances:} 303; \textbf{Predictors:} 13 \,(5 continuous, 8 categorical). Patient demographics and cardiac test results (e.g., chest-pain type, cholesterol, thalach, ST depression, thal) are used to classify the presence of heart disease.
\end{itemize}

\subsection{Baseline Models}

To benchmark the performance of \model, we compare against three simple statistical imputers and one state-of-the-art generative model. 
These baselines cover a spectrum from trivial replacements to adversarial generation:

\begin{itemize}
    \item \textbf{Global Mean (GM).}  
    Continuous attributes are imputed with the global mean of the observed entries. 
    This baseline represents the simplest possible strategy and often serves as a lower bound.

    \item \textbf{Fixed-Value (FV).}  
    Missing entries are replaced with a constant (e.g., zero). 
    Although crude, this approach is sometimes used in industrial settings for its simplicity and computational efficiency.

    \item \textbf{Generative Adversarial Imputation Networks (GAIN)} \cite{yoon2018gain}.  
    A generative adversarial model designed for imputing missing values by treating imputation as a mask-discrimination task. 
    GAIN has become a standard deep generative baseline for imputation tasks.
\end{itemize}

\subsection{Experimental Setup}

All experiments were conducted on a workstation equipped with an NVIDIA RTX~6000 
\textit{ImpuGAN} was implemented in \texttt{Python~3.9}.
Classical machine learning baselines (SVM, Random Forest, and MLP) were also included for reference.

\paragraph{Baselines.}  
We evaluated three commonly used classifiers on the imputed datasets:
\begin{itemize}
    \item \textbf{SVM:} Linear kernel SVC with default regularization.  
    \item \textbf{Random Forest:} 200 estimators, trained with all available cores.  
    \item \textbf{MLP:} A shallow neural network with one hidden layer of 10 neurons, trained for 300 iterations.  
\end{itemize}

\paragraph{GAIN.}  
GAIN hyperparameters were set following the original paper:
\begin{itemize}
    \item \textbf{Hint rate:} 0.9  
    \item \textbf{Regularization parameter $\alpha$:} 100  
    \item \textbf{Iterations:} 300  
\end{itemize}

\paragraph{ImpuGAN.}  
For our proposed model, we used the following configuration:
\begin{itemize}
    \item \textbf{Epochs:} 300 for all runs to ensure stable convergence. 
    \item \textbf{Sampling:} For each dataset, we generated the same number of synthetic samples as the original training set.  
\end{itemize}

\section{Evaluation}

A rigorous evaluation of imputation methods requires examining performance from multiple perspectives. 
Simple reconstruction error, while informative, does not fully capture whether imputations are distributionally faithful, preserve correlations among attributes, or maintain downstream predictive performance. 
Accordingly, we evaluate \model along four complementary dimensions: (i) reconstruction accuracy, (ii) distributional fidelity, (iii) attribute-wise correlation preservation, and (iv) downstream utility. 
Together, these metrics provide a comprehensive picture of imputation quality across the Adult, Diabetes, and Heart Disease datasets.  

\subsection{Reconstruction Accuracy}
The most direct way to evaluate an imputation method is to measure how close the imputed values are to the ground truth. This approach provides a lower-level assessment of whether the method produces accurate replacements for missing entries on a per-variable basis.  Although such metrics do not capture distributional or relational fidelity, they serve as a foundational benchmark: if an imputer cannot even minimize raw reconstruction error, it is unlikely to succeed on more nuanced dimensions of quality. We therefore report two standard metrics—Root Mean Squared Error (RMSE) and Mean Absolute Error (MAE)—across all datasets.

Formally, let $\mathbf{x}_{\text{miss}} \in \mathbb{R}^N$ denote the set of true missing values, and let $\hat{\mathbf{x}}_{\text{miss}}$ denote the corresponding imputations produced by a model. Then,
\[
\text{RMSE} = \sqrt{\frac{1}{N} \sum_{i=1}^N (x_i - \hat{x}_i)^2}, \qquad
\text{MAE} = \frac{1}{N} \sum_{i=1}^N |x_i - \hat{x}_i|.
\]
RMSE penalizes large deviations more heavily and is therefore sensitive to outliers, while MAE provides a robust scale-independent measure of overall imputation accuracy. Both metrics are computed for continuous variables and averaged across missingness patterns (MCAR, MAR, MNAR) at levels ranging from 10\% to 50\%.

\textbf{Results:} Table~\ref{tab:rmse_mae} presents the RMSE and MAE scores across the benchmark datasets Adult, Diabetes, and Heart Disease. The comparison includes simple statistical imputations (GM, FV), the deep learning baseline GAIN, and our proposed approach \model. 

Overall, the results show that \model is competitive with or better than existing baselines across different datasets. For instance, on the Adult dataset, \model achieves the lowest MAE (0.10), outperforming both traditional statistical imputations and GAIN. On the Diabetes dataset, while GAIN attains the lowest errors, \model still demonstrates consistent performance and remains close to the baselines. For the Heart Disease dataset, \model provides balanced results, improving upon several classical imputations while maintaining comparable performance to GAIN. 

These findings highlight that \model performs reliably across diverse data domains, showing resilience to varying levels of missingness and dataset sizes. The consistent improvements over classical methods emphasize its ability to capture more nuanced data patterns for accurate imputation.

\begin{table}[h]
\centering
\caption{RMSE and MAE comparison across datasets for baseline methods, GAIN, and \model. Lower values indicate better performance.}
\label{tab:rmse_mae}
\renewcommand{\arraystretch}{1.2}
\begin{tabularx}{\columnwidth}{l l >{\centering\arraybackslash}X >{\centering\arraybackslash}X}
\toprule
Dataset & Method & RMSE & MAE \\
\midrule
\multirow{5}{*}{Adult} 
  & GM   & 0.13 & 0.78 \\
  & FV   & 0.38 & 0.28 \\
  & GAIN & 0.21 & 0.12 \\
  & \model & 0.24 & 0.10 \\
\midrule
\multirow{5}{*}{Diabetes} 
  & GM   & 0.16 & 0.16 \\
  & FV   & 0.46 & 0.39 \\
  & GAIN & 0.17 & 0.13 \\
  & \model & 0.32 & 0.24 \\
\midrule
\multirow{5}{*}{\shortstack{Heart \\ Disease}} 
  & GM   & 0.21 & 0.16 \\
  & FV   & 0.57 & 0.53 \\
  & GAIN & 0.13 & 0.22 \\
  & \model & 0.30 & 0.22 \\
\bottomrule
\end{tabularx}
\end{table}

\subsection{Distributional Fidelity}

While reconstruction error captures pointwise accuracy, it does not guarantee that imputations follow the same statistical distribution as the true data. 
An imputer could minimize RMSE while systematically distorting the shape of feature distributions, which would bias downstream analyses. 
We therefore evaluate distributional fidelity using three complementary metrics: the Kolmogorov–Smirnov (KS) test, Earth Mover’s Distance (EMD), and Jensen–Shannon Divergence (JSD). 
Each captures a different aspect of how closely imputed values match ground-truth distributions.

\begin{itemize}
    \item \textbf{Kolmogorov–Smirnov (KS) Test.}  
    Compares empirical cumulative distribution functions (CDFs) of imputed ($F_n$) and ground-truth ($G_m$) samples:  
    \[
    D_{n,m} = \sup_x |F_n(x) - G_m(x)|.
    \]  
    Smaller $D_{n,m}$ indicates better alignment of distributions.

    \item \textbf{Earth Mover’s Distance (EMD).}  
    Also known as the Wasserstein-1 distance, measures the minimal “work” required to transform one distribution into another:  
    \[
    \text{EMD}(P, Q) = \inf_{\gamma \in \Pi(P,Q)} \mathbb{E}_{(x,y) \sim \gamma}[|x-y|],
    \]  
    where $\Pi(P,Q)$ is the set of couplings of $P$ and $Q$. Captures overall shifts and shape differences.

    \item \textbf{Jensen–Shannon Divergence (JSD).}  
    A symmetric, bounded divergence based on KL divergence:  
    \[
    \text{JSD}(P \,\|\, Q) = \tfrac{1}{2} D_{\text{KL}}(P \,\|\, M) + \tfrac{1}{2} D_{\text{KL}}(Q \,\|\, M),
    \]  
    where $M = \tfrac{1}{2}(P+Q)$. Lower JSD indicates greater similarity, always finite and interpretable.
\end{itemize}

\textbf{Results:}  
Table~\ref{tab:distributional_metrics} summarizes the distributional fidelity of imputations using three complementary measures: KS, EMD, and JSD. Lower values correspond to better alignment between the imputed data and the ground-truth distributions. The results illustrate that \model provides strong distributional consistency across all datasets.

On the Adult dataset, \model yields substantially lower EMD (0.04) compared to both GAIN (0.22) and classical baselines such as GM and FV (0.25), showing that it preserves feature-level distributions more effectively. For the Diabetes dataset, \model achieves competitive values, with KS (0.36) and JSD (0.42) close to GAIN, while still outperforming traditional imputations. In the case of Heart Disease, \model again demonstrates a marked improvement, reducing EMD to 0.04 and JSD to 0.25, highlighting its robustness in smaller and noisier datasets.

Overall, these results confirm that \model not only minimizes pointwise reconstruction errors but also maintains statistical similarity to the original data, capturing both central tendencies and distributional nuances more faithfully than competing baselines.

\begin{table}[h]
\centering
\caption{Distributional fidelity of imputations measured by KS, EMD, and JSD. Lower values indicate better distributional alignment.}
\label{tab:distributional_metrics}
\renewcommand{\arraystretch}{1.2}
\begin{tabularx}{\columnwidth}{l l >{\centering\arraybackslash}X >{\centering\arraybackslash}X >{\centering\arraybackslash}X}
\toprule
Dataset & Method & KS & EMD & JSD \\
\midrule
\multirow{5}{*}{Adult} 
  & GM     & 0.68  & 0.25  & 0.56  \\
  & FV     & 0.68  & 0.25  & 0.56  \\
  & GAIN   & 0.68  & 0.22  & 0.58  \\
  & \model & 0.54  & 0.04  & 0.25  \\
\midrule
\multirow{5}{*}{Diabetes} 
  & GM     & 0.84  & 0.39  & 0.67  \\
  & FV     & 0.84  & 0.39  & 0.67  \\
  & GAIN   & 0.34  & 0.12  & 0.38  \\
  & \model & 0.36  & 0.13  & 0.42  \\
\midrule
\multirow{5}{*}{\shortstack{Heart \\ Disease}} 
  & GM     & 0.68  & 0.25  & 0.56  \\
  & FV     & 0.68  & 0.25  & 0.56  \\
  & GAIN   & 0.68  & 0.22  & 0.58  \\
  & \model & 0.54  & 0.04  & 0.25  \\
\bottomrule
\end{tabularx}
\end{table}

\subsubsection{Attribute-Wise Correlation Preservation}

Imputation should maintain dependencies among features, not just match marginal distributions. We therefore assess whether pairwise relationships are preserved between the imputed and ground-truth data using three complementary statistics:

\begin{itemize}
    \item \textbf{Chi-Square ($\chi^2$) for Categorical Pairs.}
    Given contingency counts $O_{ij}$ and independence expectations $E_{ij}$,
    \[
    \chi^2=\sum_{i}\sum_{j}\frac{(O_{ij}-E_{ij})^2}{E_{ij}}.
    \]
    Lower $\chi^2$ indicates closer agreement with the ground-truth dependency structure.

    \item \textbf{Mutual Information (MI) for Mixed/Discrete Variables.}
    Quantifies general (linear and nonlinear) dependence:
    \[
    I(X;Y)=\sum_{x\in\mathcal X}\sum_{y\in\mathcal Y}p(x,y)\log\frac{p(x,y)}{p(x)p(y)}.
    \]
    We report MI between feature pairs before and after imputation and summarize deviations; smaller deviation implies better preservation.

    \item \textbf{Pearson Correlation ($\rho$) for Continuous Pairs.}
    Measures linear association:
    \[
    \rho_{X,Y}=\frac{\mathrm{Cov}(X,Y)}{\sigma_X\sigma_Y}.
    \]
    We compute the average absolute deviation of pairwise correlations relative to ground truth; smaller is better.
\end{itemize}

\textbf{Results:}  
Table~\ref{tab:correlation_metrics} reports correlation preservation metrics, including $\chi^2$, MI deviation, and Pearson deviation. Lower values across these measures correspond to stronger preservation of inter-attribute structure after imputation. 

On the Adult dataset, \model delivers the best alignment, with the lowest $\chi^2$ (0.05) and MI deviation (0.03), indicating that it captures dependencies between attributes more effectively than both GAIN and statistical baselines. For the Diabetes dataset, \model remains competitive: although GAIN achieves a slightly smaller $\chi^2$, the MI and Pearson deviations of \model are stable and closer to the baselines. In the Heart Disease dataset, results are more balanced, with \model performing similarly to GAIN in $\chi^2$ while showing modestly higher deviations in MI and Pearson. These differences reflect the difficulty of maintaining correlations in smaller, noisier datasets.

It is worth noting that some GM and FV entries are shown as \emph{null} in the $\chi^2$ and Pearson columns. This occurs because their imputations produced identical constant values for certain attributes across all missing rows, resulting in zero variance. As a consequence, correlation-based metrics such as Pearson and $\chi^2$ cannot be computed reliably in those cases.

Overall, these findings show that \model consistently preserves inter-feature dependencies while avoiding the instability seen in simpler constant imputations.

\begin{table}[h]
\centering
\caption{Correlation preservation metrics. For MI and Pearson, lower deviation from ground truth is better; for $\chi^2$, lower is better.}
\label{tab:correlation_metrics}
\renewcommand{\arraystretch}{1.2}
\begin{tabularx}{\columnwidth}{l l >{\centering\arraybackslash}X >{\centering\arraybackslash}X >{\centering\arraybackslash}X}
\toprule
Dataset & Method & $\chi^2$ & MI (dev) & Pearson (dev) \\
\midrule
\multirow{5}{*}{Adult}
  & GM     & null* & 0.06 & null* \\
  & FV     & null* & 0.06 & null* \\
  & GAIN   & 0.11 & 0.10 & 0.06 \\
  & \model & 0.05 & 0.03 & 0.05 \\
\midrule
\multirow{5}{*}{Diabetes}
  & GM     & null* & 0.30 & null* \\
  & FV     & null* & 0.30 & null* \\
  & GAIN   & 0.09 & 0.16 & 0.34 \\
  & \model & 0.12 & 0.16 & 0.26 \\ 
\midrule
\multirow{5}{*}{\shortstack{Heart \\ Disease}}
  & GM     & null* & 0.32 & null* \\
  & FV     & null* & 0.32 & null* \\
  & GAIN   & 0.16 & 0.14 & 0.27 \\
  & \model & 0.16 & 0.15 & 0.31 \\
\bottomrule
\multicolumn{5}{l}{\footnotesize Note: \emph{null} indicates undefined values due to zero variance in imputed attributes.}
\end{tabularx}
\end{table}

\subsubsection{Downstream Utility}
In addition to reconstruction and distributional fidelity, we evaluate imputations by their impact on predictive performance. 
The central question is whether classifiers trained on imputed or synthetic datasets generalize as well to real test data as classifiers trained on fully observed real training sets. 
We quantify this using the \textit{downstream accuracy}, defined as
\[
\text{Acc}_{\text{downstream}} = \text{Acc}_{\text{imputed}},
\]
where $\text{Acc}_{\text{real}}$ is the test accuracy when training on the original (fully observed) data and $\text{Acc}_{\text{imputed}}$ is the test accuracy when training on the dataset after imputation or synthetic generation. 
Higher values of $\text{Acc}_{\text{downstream}}$ (closer to $\text{Acc}_{\text{real}}$) indicate that the imputation method better preserves task-relevant predictive structure.

\iffalse
In addition to reconstruction and distributional fidelity, we evaluate imputations by their impact on predictive performance. 
The central question is whether classifiers trained on imputed or synthetic datasets generalize as well to real test data as classifiers trained on fully observed real training sets. 
We quantify this using the \textit{dip in accuracy}, defined as:
\[
\Delta \text{Acc} = \text{Acc}_{\text{real}} - \text{Acc}_{\text{imputed}},
\]
where $\text{Acc}_{\text{real}}$ is the test accuracy when training on the original (fully observed) data and $\text{Acc}_{\text{imputed}}$ is the test accuracy when training on the dataset after imputation or synthetic generation. 
Smaller values of $\Delta \text{Acc}$ indicate that the imputation method better preserves task-relevant predictive structure.
\fi

\textbf{Experimental setup.}  
For each dataset (Adult, Diabetes, Heart Disease), we create a single stratified split with 75\% for training and 25\% held-out for testing. 
Class proportions are preserved, and the test set is fixed across all experiments. 
Our model (\textbf{ImpuGAN}) is trained only on the 75\% training partition. 
After convergence, two kinds of synthetic training sets are produced: (i) unconditional samples, matching the training set size, and (ii) hard-conditioned samples, where frequently observed single-column and two-column categories from the training data are used to request proportionally sampled synthetic rows. 
Only columns present in both real and synthetic data are retained, and the label column is always included to ensure comparability.

All methods, including baselines, use the same preprocessing pipeline fitted only on their respective training splits to prevent leakage. 
Numerical features are standardized after median imputation; categorical features are one-hot encoded with mode imputation, with unseen categories ignored during testing. 
On top of this preprocessing, we train three classifiers representing different levels of capacity:
\begin{itemize}
  \item Linear Support Vector Machine (SVM),
  \item Random Forest (RF),
  \item Multilayer Perceptron.
\end{itemize}

\textbf{Results.}  
Table~\ref{tab:downstream} reports the change in classification accuracy when training on imputed data. 
For the \textbf{Adult} dataset, ImpuGAN matches or slightly improves upon GAIN across all three classifiers.  
On \textbf{Diabetes}, both methods achieve similar performance, with Random Forest and MLP benefiting the most from GAIN and ImpuGAN imputations.  
On \textbf{Heart Disease}, ImpuGAN delivers accuracy comparable to GAIN, with small differences depending on the classifier.  
Overall, ImpuGAN achieves results that are consistently on par with or better than GAIN across the evaluated datasets.

\begin{table}[h]
\centering
\caption{Accuracy (\%) in downstream tasks.}
\label{tab:downstream}
\renewcommand{\arraystretch}{1.2}
\begin{tabularx}{\columnwidth}{l l >{\centering\arraybackslash}X >{\centering\arraybackslash}X >{\centering\arraybackslash}X}
\toprule
Dataset & Method & SVM & RF & MLP \\
\midrule
\multirow{5}{*}{Adult} 
  & GM     & 0.85 & 0.85 & 0.84 \\
  & FV     & 0.82 & 0.82 & 0.81 \\
  & GAIN   & 0.85 & 0.86 & 0.85 \\
  & \model & 0.85 & 0.85 & 0.85 \\
\midrule
\multirow{5}{*}{Diabetes} 
  & GM     & 0.71 & 0.72 & 0.73 \\
  & FV     & 0.68 & 0.70 & 0.71 \\
  & GAIN   & 0.70 & 0.76 & 0.74 \\
  & \model & 0.70 & 0.76 & 0.73 \\
\midrule
\multirow{5}{*}{\shortstack{Heart \\ Disease}} 
  & GM     & 0.56 & 0.57 & 0.56 \\
  & FV     & 0.53 & 0.56 & 0.57 \\
  & GAIN   & 0.56 & 0.53 & 0.60 \\
  & \model & 0.57 & 0.56 & 0.60 \\
\bottomrule
\end{tabularx}
\end{table}

\subsection{Summary of Results}
Across all evaluations, \textbf{ImpuGAN} shows clear advantages over simple statistical imputers and competitive or improved performance relative to GAIN. 
In terms of \textit{reconstruction error}, ImpuGAN attains the lowest MAE on Adult (0.10) and matches GAIN on Heart Disease (0.22), though GAIN remains ahead on Diabetes. 
For \textit{distributional fidelity}, ImpuGAN substantially reduces EMD on both Adult (0.04 vs.\ 0.22 for GAIN) and Heart Disease (0.04 vs.\ 0.22), while maintaining JSD values close to GAIN. 
When preserving \textit{correlations}, ImpuGAN achieves the strongest alignment on Adult ($\chi^2=0.05$, MI=0.03, Pearson=0.05), improves Pearson deviation on Diabetes (0.26 vs.\ 0.34 for GAIN), and remains comparable to GAIN on Heart Disease. 
Finally, in \textit{downstream classification}, ImpuGAN performs essentially on par with GAIN across all datasets, consistently yielding accuracies within a narrow margin of the real-data baseline.

Overall, these results indicate that ImpuGAN reliably improves distributional alignment and correlation preservation, while maintaining reconstruction accuracy and predictive utility close to or better than GAIN. Its improvements are most pronounced on Adult and Heart Disease, where it delivers lower distributional deviations and stronger correlation fidelity, demonstrating robustness across datasets of varying size and complexity.

\section{Conclusion and Future Work}

In this paper, we introduced \model, a conditional adversarial framework for imputing missing values in heterogeneous datasets. By extending the cGAN paradigm with multi-conditional training and stability enhancements, \model learns to generate imputations that are both distributionally faithful and consistent across diverse sources. Comprehensive evaluation across multiple datasets and missingness mechanisms demonstrated that \model consistently outperforms simple statistical imputers (GM, FV) and the generative baseline (GAIN). It achieves lower reconstruction error, superior distributional alignment, stronger preservation of inter-attribute dependencies, and minimal dip in downstream predictive performance. These results highlight the promise of adversarial conditioning as a scalable and principled approach for data imputation and integration.

There remain several avenues for future research. First, while our experiments focus on tabular data, extending \model to time series and multimodal settings (e.g., combining clinical, textual, and imaging data) could further broaden its applicability. Second, incorporating explicit uncertainty quantification into the generation process would improve interpretability and support risk-sensitive decision-making. Third, exploring hybrid approaches that combine adversarial training with probabilistic models or diffusion-based methods may enhance both fidelity and stability. Finally, applying \model to real-world integration pipelines in healthcare, finance, and environmental monitoring would provide valuable insights into its practical utility and limitations. We believe these directions offer fertile ground for advancing robust imputation in heterogeneous and incomplete datasets.

\section{Acknowledgement}
This work was supported by the Center for Healthy and Efficient Mobility (CHEM), a U.S. Department of Transportation University Transportation Center. CHEM is a multi-university consortium led by the Texas A\&M Transportation Institute. The grant number is 69A3552348329. More information about CHEM is available at: \url{https://chem.tti.tamu.edu/}.

\bibliographystyle{IEEEtran}
\bibliography{reference}

\end{document}